\documentclass[10pt,twocolumn,letterpaper]{article}

\usepackage{cvpr}
\usepackage{times}
\usepackage{epsfig}
\usepackage{graphicx}
\usepackage{amsmath}
\usepackage{amssymb}
\usepackage{multirow}
\usepackage{algorithm}
\usepackage{subcaption}
\usepackage{algpseudocode}
\usepackage{scrextend}
\usepackage{tabularx}
\usepackage{booktabs}
\usepackage{bbm}
\usepackage{breqn}
\usepackage{verbatim}
\usepackage{eso-pic}
\pdfoutput=1

\usepackage[pagebackref=true,breaklinks=true,letterpaper=true,colorlinks,bookmarks=false]{hyperref}

\cvprfinalcopy 

\def\cvprPaperID{****} 

\ifcvprfinal\fi
\begin{document}

\title{Personalized Automatic Estimation of Self-reported Pain Intensity\\ from Facial Expressions}

\author{Daniel Lopez Martinez$^{1,2}$, Ognjen (Oggi) Rudovic$^2$ and Rosalind Picard$^2$\\\\
$^1$ Harvard-MIT Division of Health Sciences and Technology, Massachusetts Institute of Technology\\ $^2$ MIT Media Lab, Massachusetts Institute of Technology, Cambridge, MA, USA\\ 
{\tt\small \{dlmocdm,orudovic,picard\}@media.mit.edu}
}

\maketitle

\begin{abstract}
Pain is a personal, subjective experience that is commonly evaluated through visual analog scales (VAS). While this is often convenient and useful, automatic pain detection systems can reduce pain score acquisition efforts in large-scale studies by estimating it directly from the participants' facial expressions. In this paper, we propose a novel two-stage learning approach for VAS estimation: first, our algorithm employs Recurrent Neural Networks (RNNs) to automatically estimate Prkachin and Solomon Pain Intensity (PSPI) levels from face images. The estimated scores are then fed into the {\it personalized} Hidden Conditional Random Fields (HCRFs), used to estimate the VAS, provided by each person. Personalization of the model is performed using a newly introduced facial expressiveness score, unique for each person. To the best of our knowledge, this is the first approach to automatically estimate VAS from face images. We show the benefits of the proposed personalized over traditional non-personalized approach on a benchmark dataset for pain analysis from face images. 
\end{abstract}

\section{Introduction}

Pain is a distressing experience associated with actual or potential tissue damage with sensory, emotional, cognitive and social components \cite{DeCWilliams2016}. Accurate, valid, and reliable measurement of pain is essential for the  diagnosis and management of pain in the hospital and at home, and to ensure accurate evaluation of the relative effectiveness of different therapies. 
In order to develop  automatic methods of objectively quantifying an individuals experience of pain, several physiologic variables have been measured, such as skin conductance  and heart rate \cite{Treister2012}. In general, however, these markers do not correlate strongly enough with pain to warrant their use as a surrogate measure of pain \cite{Younger2009}.
In the absence of a valid and reliable objective, physiologic marker of pain, patient's self-report provides the most valid measure of this fundamentally subjective experience \cite{textbookPain21}.
Various pain rating scales have been developed to capture patient's self-report of pain intensity \cite{Hawker2011,Haefeli2006}. The Visual Analogue Scale (VAS), Numerical Rating Scale (NRS),
Verbal Rating Scale (VRS), and Faces Pain Scale-Revised (FPS-R) are among the most commonly used scales. While evidence supports the reliability and validity of each of these measures across many populations \cite{Ferreira-Valente2011}, each measure has strengths and weaknesses that make them more appropriate for different applications \cite{Williamson2005}.
In the research setting, for example, VAS is usually preferred since it is statistically the most robust as it can provide ratio level data \cite{Williamson2005,Price1983},  allowing parametric tests to be used in statistical analysis. Furthermore, VAS is  the most common pain intensity scale in clinical trials \cite{Todd1996,Farrar2000,Moore1997,Jensen2003,Aicher2012,Jensen2005}. 
When using the VAS, patients are asked to mark the pain that they are experiencing on a 10cm-long horizontal line labeled "no pain" on the far left and "worst pain ever" on the far right. Pain intensity is determined by the length of the line as measured from the left-hand side to the point marked  \cite{Mohan2010,Hawker2011}. The following cut points on the pain VAS are recommended: no pain (0-0.4 cm), mild pain (0.5-4.4 cm), moderate pain (4.5-7.4 cm), and severe pain (7.5-10 cm) \cite{Hawker2011}.
Unfortunately, VAS is highly impractical and inefficient to measure and lacks utility in key patient populations, as it requires patients to have intact fine motor skills that may be limited by  illness or injury.
 
To circumvent the challenge of acquiring patients' self-reported pain scores, facial indicators of pain \cite{Williams2002a} have been used in  automatic pain detection systems. However, all existing studies have focused on the estimation of the Prkachin and Solomon Pain Intensity (PSPI) score, a metric that measures pain as a linear combination of the intensities of facial action units (see Sec.\ref{eq:ifes}) \cite{Prkachin2008}, instead of VAS. While the facial response to pain certainly consists of a core set of facial actions, the relationship between facial cues and VAS scores is not clear \cite{Prkachin2008}. For example, it has been shown that men and women show different levels of pain expression for the same stimulus \cite{Sullivan2000}. In fact, many studies have found low correlations between facial expression and self-reported pain \cite{Prkachin2008}. Therefore, an approach to automatically estimate self-reported pain intensities from facial cues should also account for individual's differences in facial expressiveness of pain. 

This work addresses the challenges of estimating VAS scores from facial cues. To this end, we propose a hierarchical learning framework that exploits the modeling power of sequence classification models. Specifically, we first employ a bidirectional long short-term memory recurrent neural net (LSTM-RNN) \cite{Hochreiter1997} to estimate PSPI from facial landmarks extracted from face images. Then,  PSPI is used as input to the hidden conditional random field (HCRF) \cite{Quattoni2004,Wang2006a} to estimate VAS of target person. The key to our approach is the personalization of the target classifier via the newly introduced individual facial expressiveness score (I-FES). Specifically, I-FES quantifies the relative disagreement between an externally observed pain intensity (OPI) rating and the patient's self-reported VAS. By using I-FES to augment the PSPI input to HCRF, we account for person-specific biases in the VAS rating of pain levels. To the best of our knowledge, this is the first approach to automatically estimate VAS from face images. We show on the UNBC-McMaster Shoulder Pain Expression Archive Database \cite{unbc} that the proposed personalized approach for automatic VAS estimation outperforms largely traditional (unpersonalized) approaches.


\section{Related Work}

Although there is much research in the automated recognition of affect from facial expression (for surveys see: \cite{Calvo2010,zeng2009survey}),  until recently only a handful of works have focused on  automated pain estimation. Due to rapid advances in computer vision and also the recent release of the UNBC-McMaster dataset \cite{unbc}, pain analysis from face images have seen significant advances. Specifically, this dataset provides videos with each frame coded in terms of Prkachin and Solomon Pain Intensity (PSPI) \cite{Prkachin2008} score, defined on an ordinal scale 0-15. This is considered to be an objective pain score, in contrast to subjective pain ratings such as VAS. Despite the fact that VAS is still the most commonly accepted pain score in clinical settings, all existing automatic methods for pain estimation from pain images focused on prediction of PSPI scores. We outline below some of the recently published works.


Face shape features have been used in artificial neural networks to classify images of persons' faces in a typical mood versus a pain inducing task \cite{Monwar2006}. Likewise, \cite{Lucey2011,Ashraf2009a} used Active Appearance Models (AAM) - based features combined with Support Vector Machine (SVM) classifiers to classify pain versus no pain images. On the other hand, instead of treating pain as binary (pain - no pain), \cite{Hammal2012a} attempted estimation of pain intensity on a 4 level scale, using one-versus-all SVM classifiers. \cite{Kaltwang2012b} performed estimation of the full pain range (0-15), using a framework of Relevance Vector Regression models (RVR). Recently, \cite{Rodriguez2017} have directly exploited the temporal axis information by using long short-term memory (LSTM) neural networks. In their approach, raw images are fed to a convolutional neural network that extracts frame features. These features are subsequently used by the LSTM to predict a PSPI score for each frame. Therefore, this deep learning approach is able to leverage the temporal information associated with facial expressions. \cite{Egede2017}  also used a deep learning approach to perform automatic continuous PSPI prediction using a combination of hand-crafted and deep-learned features. 
 
Due to subtle changes in facial expressions and inter-subject variability, per frame PSPI estimation is very challenging. Hence, some of the works attempted pain intensity estimation per sequence. For instance, \cite{Sikka2013} proposed a multi-instance learning framework to identify the most expressive (in terms of pain) segments within image sequences. Likewise, \cite{Rusk2016} proposed a HCRF framework for semi-supervised learning of the peak of the pain in image sequences, using weakly labeled pain images. However, these per sequence labels are derived using heuristics, and do not relate to established pain ratings, such as VAS scores.

Aside from the UNBC-McMaster dataset \cite{unbc}, there have been a variety of pain recognition studies based on other datasets (e.g., see \cite{werner2012cl,Wilkie95}). Note also that  pain detection has been attempted from other modalities including upper body movements combined with face \cite{Gunes2009,Joshi2013}, physiological signals such as skin conductance and heart rate \cite{Treister2012}, and brain hemodynamic responses using NIRS \cite{Aasted2016, Yucel2015} and fMRI \cite{Mackey2013, Wager2013}. Nevertheless, none of these works attempted automatic estimation of VAS, and in a personalized manner.

\section{Personalized RNN-HCRFs: The Model}
\label{pmodel}
\begin{figure}
	\centering
	\includegraphics[width=0.40\textwidth]{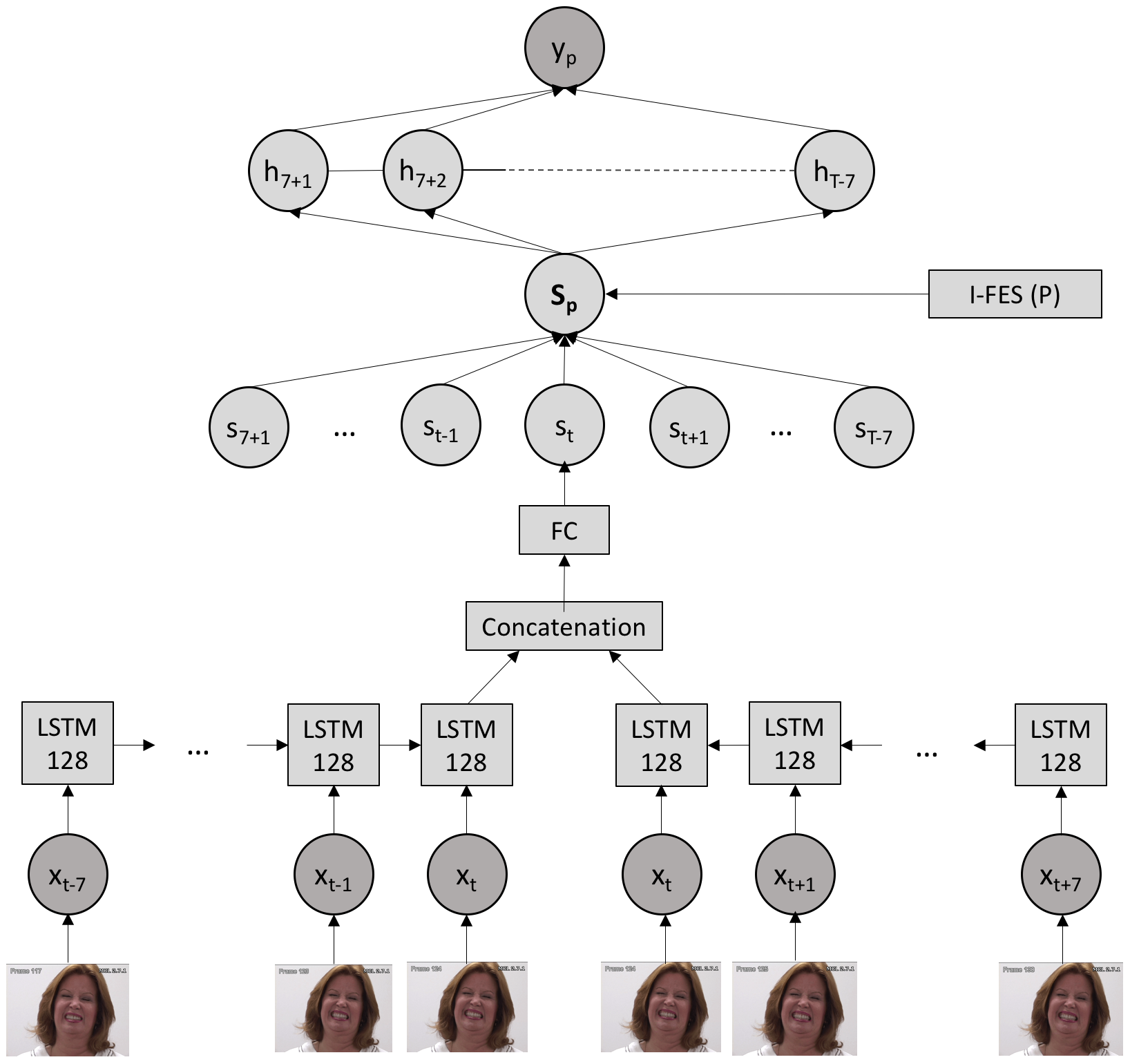}
    \caption{Graphical representation of the proposed pRNN-HCRF for estimation of VAS ($y_p$) from facial landmarks ($x$) at each frame in a sequence. The bidirectional LSTM-RNN and HCRF layers are connected  via the estimated PSPI ($s$) and individual facial expressiveness ratio (I-FES). The latter is used to personalize the model parameters.}
    \label{fig:model}
\end{figure} 


{\bf Notation.}\label{notation} We consider a sequence classification problem, where we are given $N_i$ image sequences of person $i=1,\dots,L$, and $L$ is the number of target persons. The sequences of each person are annotated in terms of VAS as $\mathcal{V}=\{V_{1},\dots,V_{L}\}$, where $V_{i}=\{v_{i}^{1},\dots,v_{i}^{N}\}$ and individual per-sequence VAS scores are $v_i\in \{0,...,10\}$. Likewise, for observed pain intensity (OPI) scores we have: $\mathcal{O}=\{O_{1},\dots,O_{L}\}$, $O_{i}=\{o_{i}^{1},\dots,o_{i}^{N}\}$ and $o_i\in \{0,...,5\}$. Furthermore, each sequence is represented by a set of input features (in our case, locations of facial landmarks -- see Fig.\ref{fig:distributions}C) and the corresponding PSPI scores, given as pairs ($X,S$), respectively. Here, $X=\{x_1, \dots,x_T\}$ and $S=\{s_1,\dots,s_T\}$, and duration $T$ can vary for each sequence. The pairs $\{x_t,s_t\}_{t=1}^T$ are assumed to be i.i.d.~samples from an underlying but unknown distribution, and $x_t\in \mathcal{R}^D$, where $D$ is the input dimension, and $s_t\in \{0,...,15\}$. We exploit all this information (as described below) to define and learn our model. 

We exploit the inherent hierarchical nature of the target problem to define our model for personalized VAS score prediction from image sequences. To this end, we first leverage the modeling power of Long Short-Term Memory Recurrent Neural Networks (LSTM-RNNs)\cite{Hochreiter1997} to estimate (in an unpersonalized manner) the PSPI scores from input features $X$ (Sec.\ref{rnns}). These are then endowed with the newly introduced Individual Facial Expressiveness Score (I-FES) denoted by $p_i$ (Sec.\ref{sec:ifes}), and used as input to the top layer of our model, based on Hidden Conditional Random Fields (HCRF) \cite{Quattoni2004,Wang2006a}, that performs sequence classification in terms of target VAS scores $v_i$. To account for the "personalized dynamics", we use the notion of latent states of HCRF, denoted as $H=\{h_1,\dots,h_{T}\}$ (Sec.\ref{hcrf}). The graph model of the proposed personalized RNN-HCRF (pRNN-HCRF) is depicted in Fig.\ref{fig:model}.

\subsection{RNNs for PSPI Estimation}
\label{rnns}
The first step in our approach is to obtain an objective estimate of the pain intensity, as encoded by the (manually coded) PSPI \cite{unbc}. To this end, various approaches based on static classifiers/regressors (e.g., SVMs, SVRs) have been proposed \cite{Kaltwang2012b}. To capture temporal information within an image sequence of a person with varying pain intensity levels, dynamic classification methods based on Hidden Markov Models (HMMs) or Conditional Random Fields, and their extensions are typically used \cite{Rudovic2013}. However, these methods usually assume first order Markov dependency, thus failing to capture long-term dependencies in image sequences. 

We use the long short-term memory recurrent neural nets (LSTM-RNNs) \cite{Hochreiter1997}, as they have recently shown great success in sequence learning tasks such as speech recognition \cite{Graves2014} and sleep/wake classification \cite{weixuan2017}. In this work, we employ a bidirectional LSTM-RNNs architecture \cite{weixuan2017} (shown in Fig.\ref{fig:model}) to estimate the PSPI values from input features $X$ (i.e., a sequence of facial landmarks). While the traditional RNNs are unable to learn temporal dependencies longer than a few time steps due to the vanishing gradient problem \cite{Hochreiter2001}, LSTM-RNNs overcome this by introducing recurrently connected memory blocks instead of traditional neural network nodes. In an LSTM-RNN, each block contains one or more recurrently connected memory cells, along with three multiplicative gate units: the input, output, and forget gates \cite{Hochreiter1997}. Their role is to give access to long range context information to each block (in our case, the relationships between facial expressions across a time-window). This, in turn, allows the network to store and retrieve information over longer time windows.

To train our LSTM-RNN\footnote{Further in the text, we refer to this model as RNN.} for PSPI estimation, we consider the standard regression setting: each PSPI score at time $t$ in an image sequence is treated as a continuous value $s_t$ (PSPI $\in\{0,\dots,15\}$ scaled to $s_t=(0-1)$). The inputs are the normalized locations of 66 facial landmarks ($x_t$), provided by the database creators and depicted in Fig.\ref{fig:distributions}. Our model then tries to estimate $s_t$ for each frame $t$. To do so, the LSTM units in our RNN consider windows of 15 frames: $\{x_{t-7},...,x_t,...,x_{t+7}\}$ (see Fig.\ref{fig:model}). This has been shown to be sufficient to capture local (dynamic) changes in facial expressions due to the pain onset/offset phases \cite{Rodriguez2017}. To select the number of  units in the LSTM blocks, we performed a validation on the subset of training data. The number of units was set to 128. The output of the two LSTM-RNNs is concatenated and fed to a fully-connected layer with rectified linear units, which produces a single score for each 15-frame window. The neural network was trained using root mean square propagation with mean squared error loss. The whole algorithm was implemented using deep learning frameworks Theano 0.8.2 and Keras 1.2.1.

\subsection{Individual Facial Expressiveness Score (I-FES)}
\label{sec:ifes}
Most work on automatic pain estimation from faces has focused on estimating the PSPI, a metric that measures pain as a linear combination of the intensities of facial action units (AUs):
\begin{equation}
\textnormal{PSPI = AU4+max(AU6,AU7)+max(AU9,AU10)+AU43},
\end{equation}
where the included AU's (see Fig.\ref{fig:distributions}B) are defined on a 0-5 ordinal scale, and correspond to brow lowering (AU4), orbital tightening (AU6\&7), eye closure (AU43), nose wrinkling, and lip raise (AU9\&10) \cite{Calvo2015}. These are derived by manual coding of each face image using the Facial Action Coding System (FACS) \cite{Ekman2002a}. 

\begin{figure}
	\centering
	\includegraphics[width=0.42\textwidth]{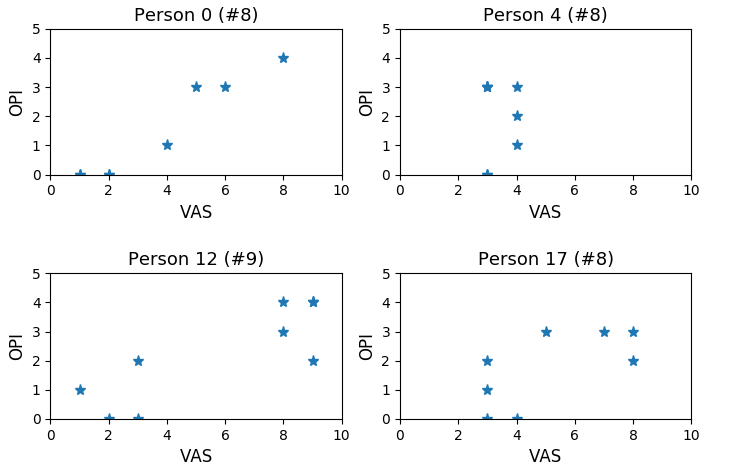}
    \caption{The VAS vs OPI scores for four different subjects (along with the number of available sequences). Note the different correlation patterns exhibited for each person: linear (top-left), no-correlation (top-right), and bi-modal (bottom row). All these pose challenges in predicting the individual VAS. For this reason, we account for these individual biases via the proposed I-FES (Eq.\ref{eq:ifes}) in our pRNN-HCRF.}
    \label{fig:oprvsvas}
\end{figure}

Estimating VAS poses a big challenge since facial expressions and self-reports are typically poorly correlated due to person-specific biases in facial expressions of pain \cite{Prkachin2008}. To account for these individual differences (e.g., see Fig.\ref{fig:oprvsvas}), we introduce the Individual Facial Expressiveness Score (I-FES), which captures the ratio of OPI ($o$), obtained by independent observers, to VAS ($v$). Specifically, for person $i$, we define I-FES ($p_i$), as: 
\begin{equation}
p_i=\left\{\begin{matrix}
\frac{1}{\alpha}\sum_{k=1}^{\alpha}\frac{o_i^k+1}{v_i^k+1},\,\, \text{iff}\,\, \alpha>0\\ 
1\,\,\,\,\,\,\,\,\,\,\,\,\,\,\,\, \text{,\,       otherwise}
\end{matrix}\right.
\label{eq:ifes}
\end{equation}
where $\alpha$ is the number of sequences used to calculate I-FES for target person. When $\alpha=0$, I-FES is set to 1 (assuming the perfect agreement), whereas $\alpha=1,..,N_{L_i}$ indicates the number of sequences of person $i$. To avoid division by 0 we add +1 in Eq.\ref{eq:ifes}. Note the intuition behind the proposed I-FES: for a given self-reported VAS, the observed pain intensity (OPI), as noted by an external observer, is expected to vary significantly from person to person depending on their level of facial expressiveness. 
Using I-FES, we quantify these relative differences in the ratings, with the OPI rater acting as the 'reference point' for all target persons.

\subsection{HCRF for Personalized VAS Estimation}
\label{hcrf}
HCRFs are a class of models for dynamic multi-class classification of sequential data \cite{Quattoni2004,Wang2006a}. They are a generalization of Conditional Random Fields (CRFs) \cite{Lafferty2001b} proposed for modeling/decoding of the state-sequence (e.g., PSPI scores for each image frame \cite{Rudovic2013}). In HCRFs, the first-order log-linear CRFs are usually employed as building blocks. This is attained by introducing a top node in the graph structure of CRFs (see Fig.\ref{fig:model}), representing the sequence label (in our case, the target VAS score), and the temporal states are treated as hidden variables in the model, and denoted by $H=\{h_1, \dots,h_{T}\}$. Formally, using the notation introduced above, HCRFs combine the score functions of $K$ CRFs, one for each class $v\in\{1,\dots, K\}$, via the following score function:
\begin{equation}
F(v,{S_p},{H};{\boldsymbol\Omega}) = \sum_{k=1}^K I(k=v) \cdot f({S_p},{H};{\theta}_k).
\label{eq:hcrf_score}
\end{equation}
We define the personalized HCRF features as $S_p=[X\,p]$, where $X$ is the sequence of input features and $p$ the I-FES defined in Sec.\ref{sec:ifes}. Furthermore, ${\boldsymbol\Omega} = \{{\theta}_k\}_{k=1}^K$ denotes the model parameters, and $f(\cdot,\cdot;{\theta}_k)$ is the $k$-th CRF score function. We assume the linear-chain graph structure  $G = (U,E)$, encoded by the sum of the unary (U) and edge (E) potentials:
\begin{eqnarray}
\sum_{r \in U} {\boldsymbol\Psi}^{(U)}({S_p},h_r)+\sum_{e=(r,z) \in E}  {\boldsymbol\Psi}^{(E)}({S_p},h_r,h_z).
\label{eq:score_node_edge}
\end{eqnarray}
 The unary potentials are defined as the linear classifier:
\begin{equation}
{\boldsymbol\Psi} _{r}^{(U)}(S_p,h_r) = \sum\limits_{c = 1}^C {{\rm{I}}(h_r  = c)\cdot {u}^{\top}_k S_p},
\label{corf:node}
\end{equation}
and the edge potentials  ${\boldsymbol\Psi}^{(E)}({ S_p},h_r,h_s)$ are defined as:
\begin{equation}
{\boldsymbol\Psi}^{(E)}({S_p},h_r,h_z)=\big [I(h_r=c \ \wedge \ h_z=l)\big ]_{C\times C} \otimes m_{k}(r,z),
\label{corf:edge}
\end{equation}
where $I(\cdot)$ is the indicator function that returns 1 (0) if the argument is true (false), and $\otimes$ is Kronecker product. The role of the edge potentials is to assure the temporal consistency of the hidden states within a sequence. The model parameters are stored as $\theta_k=\{u_k,m_k\}$. 

Using the score function defined above, the joint conditional distribution of the class and state-sequence is defined as:
\begin{equation}
P(v,{{H}}|{{S_p}}) = \frac{{\exp (F(v,{S_p},{H};{\boldsymbol\Omega}))}}{{Z({{S_p}})}}.
\label{eq:hcrf_class}
\end{equation}
The state-sequence ${H}=(h_1,\dots, h_{T})$ is unknown, and it is integrated out by directly modeling the class conditional distribution:
\begin{equation}
P(v|{S_p}) = \sum_{{H}} P(v,{H}|{S_p}) = \frac{\sum_{{H}} \exp(F(v,{S_p},{H};{\boldsymbol\Omega}))}{Z({S_p})}.
\label{eq:hcrf_marg}
\end{equation}
 
Evaluation of the class-conditional $P(v|{S_p})$ depends on the partition function $Z({{S_p}}) = \sum\limits_k {{Z_k}} ({{S_p}}) = \sum\limits_k {\sum\limits_{{H}} {\exp } } (F(k,{S_p},{H};{\boldsymbol\Omega}))$, and the class-latent joint posteriors $P(k,h_r,h_z|{ x})= P(h_r,h_z|{S_p},k) \cdot P(k|{S_p})$. Both can be computed from independent consideration of $K$ individual CRFs.

The parameter optimization in the HCRF is carried out by maximizing the (regularized) negative log-likelihood of the class conditional distribution in Eq.\ref{eq:hcrf_marg}. Formally, the regularized objective function is given by:

\begin{equation}
\label{vslinf}
    \begin{split}
        RLL({\bf{\Omega }})  =   - \sum\limits_{i = 1}^N {\log P({{\bf{v}}_i}|{{\bf{S_p^i}}};{\bf{\Omega }})} + {\lambda}||\bf{\Omega }||^2,
    \end{split}
\end{equation} 

where $N$ is the number of training sequences, $P({{\bf{v}}_i}|{{\bf{S_p}}_i};{\bf{\Omega }})$ is defined by Eq.\ref{eq:hcrf_class}, and $\lambda$ controls the regularization penalty. To estimate the model parameters, the standard quasi-Newton (such as Limited-memory BFGS) gradient descent algorithm is used. 

\subsection{Learning and Inference}
\begin{figure*}
	\centering
	\includegraphics[width=0.7\textwidth]{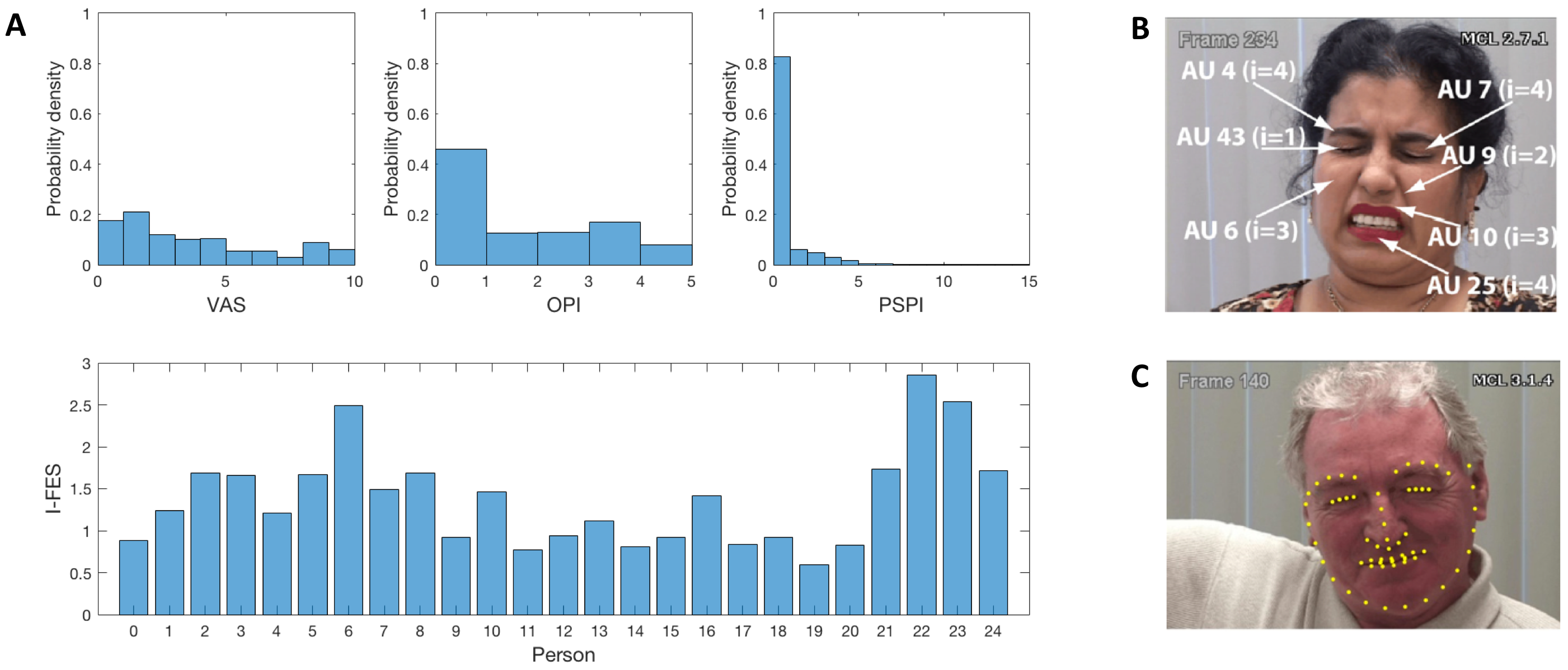}
    \caption{\textbf{(A)} Histograms of VAS, OPI and PSPI scores (top row) from the used PAIN dataset \cite{unbc}. Note that in contrast to VAS and OPI, the PSPI scores are highly biased toward low levels, which poses additional challenge in their learning. The I-FES scores (bottom row), computed using all available sequences of target persons, indicate the large individual variation in self-reported and observed pain levels. \textbf{(B)} Facial AUs and their intensity levels used to derive PSPI (=12) for target face image (taken from \cite{Hammal2012a}). \textbf{(C)} Facial landmarks ($x$) obtained using an AAM \cite{unbc}, and used as input to our approach.}
    \label{fig:distributions}
\end{figure*}
\begin{figure}
	\centering
	\includegraphics[width=0.40\textwidth]{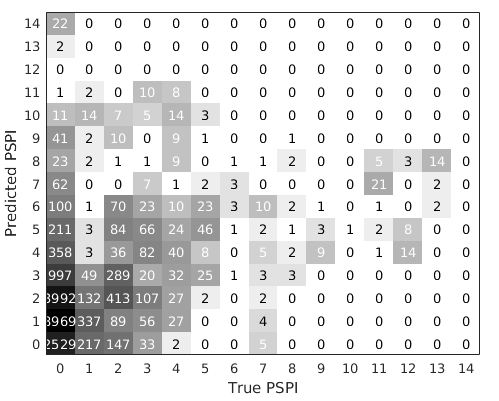}
    \caption{Confusion matrix for the predicted PSPI scores using RNN-LSTMs. Note that the model predicts well the lower intensity levels, however, it fails on high intensity levels. The latter is expected due to the highly imbalanced training PSPI data (see Fig.\ref{fig:distributions}).}
    \label{fig:confmatrixPSPI}
\end{figure}

Given facial landmarks ($X_{tr}$) extracted from image sequences of $L_{tr}$ training persons (multiple sequences per person), and the corresponding PSPI scores for each frame (stored in $S_{tr}$), we first focus on the learning of the RNNs, as described in Sec.\ref{rnns}. This step is critical as it transforms the high-dimensional input features (i.e., facial landmarks) into a low-dimensional (1-D) representation of the 'painful faces'. One can think of this as an efficient supervised dimensionality reduction approach, aimed at capturing the variance in the input that is directly related to the observed pain levels observed in face images. After  learning  the RNN model parameters ($\mathcal{Q}^{opt}$), we compute the I-FES, as denoted by $p$, for each person in the training data. Note that I-FES is unique to each person and is estimated from the ratio of VAS ($v$) and OPI ($o$) using all sequences of target persons in training set (Eq.\ref{eq:ifes}). Since we use all sequences of the training persons to estimate I-FES for each person (stored in $\mathcal{P}$)\, we  obtain a robust measure of the discrepancy between the  subjective pain rated by the person experiencing it and an external observer's (e.g., a caregiver in the clinical setting) score. 

\begin{algorithm}[t]
\small
\caption{Personalized RNN-HCRFs}
\label{alg}
\begin{algorithmic}
\\\hrulefill
\State \textbf{Learning}: Input $\mathcal{D}_{tr}=\{X_i^{tr},\mathcal{V}_i^{tr},\mathcal{O}_i^{tr}, S_i^{tr}\}_{i=1}^{L_{tr}}$\\
\textbf{Step 1:} Optimize RNNs ($\mathcal{Q}^{opt}$) given $\{S_{tr},X_{tr}\}$ \\
\textbf{Step 2:} Compute I-FES \\
\textbf{                }\textbf{for} $i=1:L_{tr}$,\,$p_i\leftarrow
\{\mathcal{V}_i^{tr},\mathcal{O}_i^{tr}\}$\,\, \textbf{end},\, $\mathcal{P}=\{p_i\}_{i=1}^{L_{tr}}$\\
\textbf{Step 3:} Optimize HCRFs ($\Omega$) \\
\textbf{                }a)\,estimate PSPI $\widetilde{S}\leftarrow$ RNN($X_{tr};{Q}^{opt}$) -- \textbf{Step 1}, $S_p=\{\widetilde{S},\mathcal{P})$\\
\textbf{                }b)\,$\underset{\bf{\Omega}^{opt}}{min}- \sum\limits_{i = 1}^{L_{tr}} {\sum\limits_{j = 1}^{N_{i}} {\log P({{\bf{v}}_i^j}|{{S_{p_i}^j}};{{\Omega }}}}) + {\lambda}||{\Omega }||^2$\,\, 
\textbf{                }\\
Output: RNNs($\cdot;\mathcal{Q}^{opt}$), HCRFs($\cdot;\Omega^{opt}$)
\\\hrulefill
\State \textbf{Inference}: \text{Input} $\mathcal{D}_{te}=\{X_{*},p_{*}\}$
\State \textbf{Step 1:} Estimate PSPI $S^*\leftarrow$ RNNs$(X^*;\mathcal{Q}^{opt})$
\State \textbf{Step 2:} Estimate VAS $v^*\leftarrow$ HCRFs$(S_p^*=\{S^*,p^*\};\Omega^{opt})$
\State \text{Output:} $S^*,v^*$
\end{algorithmic}
\end{algorithm}

The actual personalization of our approach occurs in the second stage of the learning. Namely, the estimated PSPI scores ($\widetilde{S}$), obtained using the learned RNN model, are augmented with the I-FES ($\mathcal{P}$), to obtain 2-D feature vectors for each image frame in a sequence of the target person. Such personalized feature vectors are then used as input to the HCRF model for the estimation of target VAS scores ($v$) (Sec.\ref{hcrf}). It is important to mention that while the estimated PSPI can vary from frame to frame within a sequence, the corresponding I-FES ($p$) remains constant across all sequences of a target person. In this way, it encodes the person-specific biases coming from two different rating scales (VAS\&OPI). Thus, by augmenting the person-agnostic PSPI scores ($S$), we implicitly modulate the parameters of the HCRF model. Consequently, the model adjusts the classification boundaries of the HCRF so that it fits best the VAS scores for each target person. On the other hand, since $p$ is computed using OPI as well, OPI acts as a reference point in estimating the person specific-biases in the model. We choose 11 hidden states in HCRF, the same as the number of VAS levels. The summary of our learning procedure is given in Alg.\ref{alg}. One important aspect of our model is that it separates the learning of the person-agnostic PSPI and person-specific VAS pain intensity scores. This is for the following reasons: while the first level (RNNs) focus on extracting efficient low-dimensional pain descriptor (thus, removing noise), we achieve more accurate and robust learning of target VAS scores than if the input facial landmarks are used directly in our HCRF, as confirmed by our experiments. More importantly, the contribution of the I-FES in the feature vector could easily be downplayed due to the large number of other (possibly noisy) input features (i.e., facial landmarks). We also avoid heavy parametrization of our HCRF, which can easily lead to overfitting. Lastly, by adopting the proposed two-stage approach, we allow our HCRF to focus its learning and dynamics on target person ('personalized dynamics'), rather than on filtering of noisy data of multiple persons (as done at the RNN level).

To make inference of the VAS score ($v^*$) for a (new) target person, we feed into our pRNN-HCRF the facial landmarks $x^*$ and an estimate of the I-FES ($p^*$) for that person. The output are the estimated PSPI ($s^*$) and VAS ($v^*$). Note that while this inference is straightforward, it requires, however, an estimate of the I-FES for the target person. In clinical settings, for instance, this would readily be available after several visits of the patient. In other words, we could get a few VAS/OPI scores, and use them for future predictions of the target patient's VAS. However, in the existing PAIN dataset \cite{unbc}, only several sequences per target person are available, posing a challenge for the evaluation of our approach. Therefore, to evaluate the performance of the pRNN-HCRF model, we vary the number of sequences used to compute the I-FES of target person from (0-1-2), while evaluating the model performance on all remaining sequences. Furthermore, we repeat this to avoid the bias caused by selecting specific sequences (see Sec.\ref{exp}). Note that we do not use the input features of target person, but only the ratio of the subset of VAS/OPI. Thus, we do not provide to our algorithm association between the input features and VAS scores we aim to predict.

\section{Experiments}
\label{exp}

\begin{table} \centering
\begin{tabular}{c | c | c |c}
    & RNN-LSTM & NN & SVR \\ \hline
    MAE & \textbf{0.94 (0.31)} & 1.14 (0.50) &1.16 (0.17) \\ \hline
    ICC(3,1) & \textbf{0.30 (0.10)} & 0.29 (0.17)& 0.16 (0.09) \\ \hline   
\end{tabular}
\caption{The mean MAE/ICC(3,1) and standard deviation (for 5 repetitions) for the LSTM-RNN, a neural network (NN) with one hidden layer and 200 hidden units, and SVR ($C=0.1$, $\epsilon=0.01$), for the predictions of PSPI from facial landmarks, computed on test persons.}
\label{fig:pspi_predictions}
\end{table}

\begin{table*}
\centering
\caption{The performance of different methods tested for VAS (0-10) estimation. The mean and standard deviation are computed over 5 random selection of target sequences used to compute I-FES for test persons.}
\label{my-label}
\begin{tabular}{|c|c|c|c|c}
\hline
\multirow{2}{*}{\textbf{Algorithm}} & \multirow{2}{*}{\textbf{Performance}} & \textbf{No personalization} & \multicolumn{2}{c|}{\textbf{Personalization approach}}           \\ \cline{3-5} 
                                    &                                       & \textbf{$\alpha=0$}         & \textbf{$\alpha=1$}  & \multicolumn{1}{c|}{\textbf{$\alpha=2$}}  \\ \hline
\multirow{2}{*}{HCRF(PSPI)}        & MAE                                   & 2.65                        & 2.18 (0.05)          & \multicolumn{1}{c|}{2.12 (0.23)}          \\ \cline{2-5} 
                                    & ICC                                   & 0.35                        & 0.49 (0.03)          & \multicolumn{1}{c|}{0.50 (0.13)}          \\ \hline
\multirow{2}{*}{RNN-HCRF}          & MAE                                   & 3.67                        & \textbf{2.47 (0.18)} & \multicolumn{1}{c|}{\textbf{2.46 (0.23)}} \\ \cline{2-5} 
                                    & ICC                                   & 0.04                        & \textbf{0.36 (0.08)} & \multicolumn{1}{c|}{\textbf{0.34 (0.04)}} \\ \hline
\multirow{2}{*}{SVR-HCRF}          & MAE                                   & 5.44                        & 2.98 (0.28)          & \multicolumn{1}{c|}{2.83 (0.30)}          \\ \cline{2-5} 
                                    & ICC                                   & 0.04                        & 0.22 (0.06)          & \multicolumn{1}{c|}{0.20 (0.17)}          \\ \hline
\multirow{2}{*}{HCRF$(X)$}         & MAE                                   & \textbf{2.8}                & 2.88 (0.20)          & \multicolumn{1}{c|}{2.91 (0.39)}          \\ \cline{2-5} 
                                    & ICC                                   & \textbf{0.19}               & 0.18 (0.07)          & \multicolumn{1}{c|}{0.22 (0.14)}          \\ \hline
\end{tabular}
\end{table*}

\begin{figure*}
	\centering
	\includegraphics[width=0.82\textwidth]{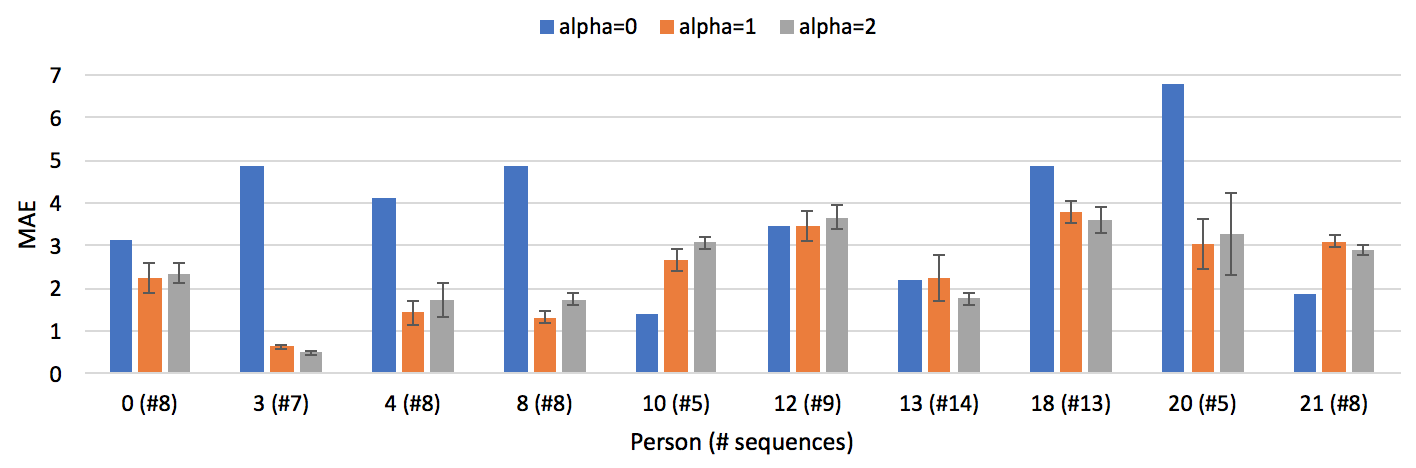}
    \caption{Mean MAE and standard deviation for the VAS estimation per test person when using pRNN-HCRF, and with varying $\alpha=0,1,2$, used to compute I-FES. We also report the number of available sequences per person. The error bars are computed over 5-random selections of target sequences used to compute I-FES for target test person.}
    \label{fig:subjects}
\end{figure*}

\begin{figure*}
	\centering
	\includegraphics[width=0.75\textwidth]{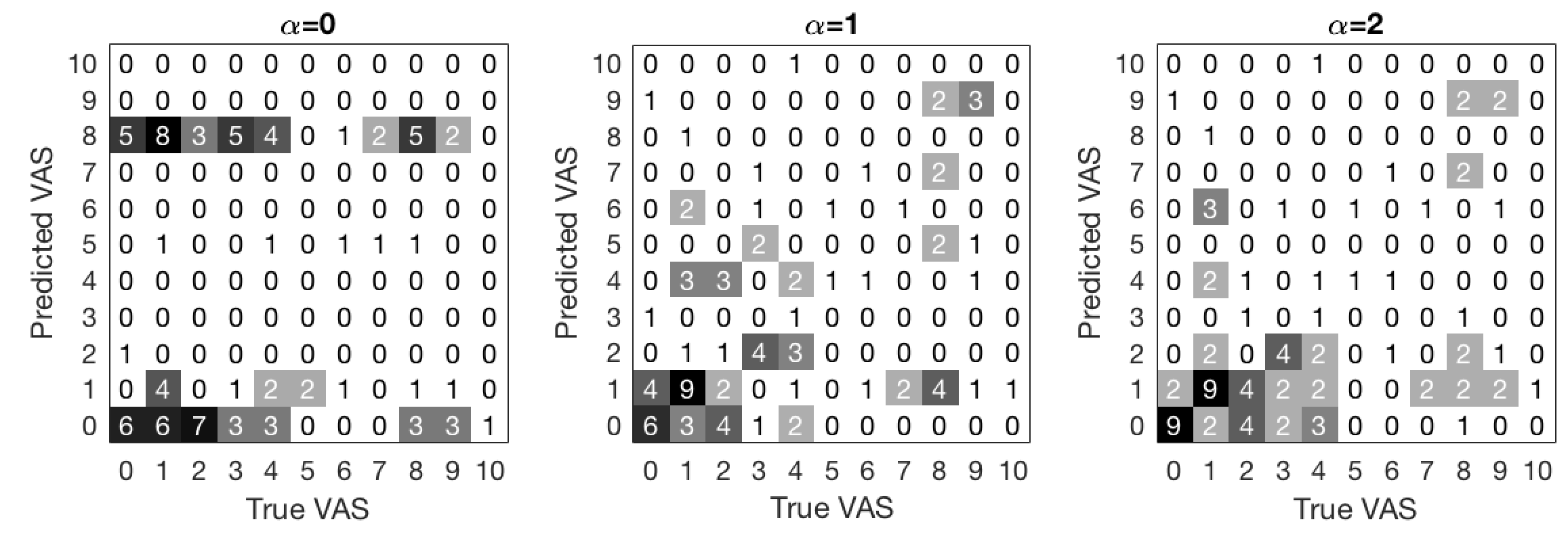}
    \caption{Confusion matrices for the predicted VAS scores using the proposed pRNN-HCRF model.}
    \label{fig:confmatrixVAS}
\end{figure*}

{\bf Dataset.} We evaluate the proposed model on the publicly available UNBC-MacMaster Shoulder Pain Expression Archive Database \cite{unbc}, which contains face videos of patients suffering from shoulder pain while performing range-of-motion tests of their arms. Two different movements are recorded: (1) the subject moves the arm himself, and (2) the subject’s arm is moved by a physiotherapist. Only one of the arms is affected by pain, but movements of the other arm are recorded as well as a control set. 200 sequences of 25 subjects were recorded (in total, 48,398 frames). Furthermore, each frame is coded in terms of the AU intensity on a six-point ordinal scale, providing the target PSPI. Also, VAS and OPI scores are provided for each image sequence (see \cite{unbc} for more details). The distributions of the coded pain levels are shown in Fig.\ref{fig:distributions}. \\\\

{\bf Features.}
As input features, we used the locations of 66 facial landmarks (see Fig.\ref{fig:distributions}), obtained by an Active Appearance Model and provided by the database creators \cite{unbc}. To reduce redundant data in the input, we applied PCA, resulting in 40D feature vectors (preserving $95\%$ of variance). Since majority of the frames are neutral (PSPI=0), for learning the PSPI estimator (the first stage), we balanced the training data so that the number of neutral frames occur with the same frequency as those with PSPI=1. We did this by removing neutral frames around the active pain segments. However, the whole sequences were used for learning VAS, and evaluation on test data. We performed the experiments in a subject independent setting, by randomly splitting the subjects in the training (15) and test (10) partitions. As evaluation measure, we used the Intra-Class Correlation ICC(3,1) \cite{Shrout1979}, which is commonly used in behavioral sciences to measure agreement between annotators (in our case, the predicted and true PSPI/VAS levels). We also report the Mean Absolute Error (MAE), commonly used for ordinal prediction tasks \cite{Kim2010,RudovicEtAlPAMI14}.

{\bf Models.} 
For comparisons with our RNN-HCRF model, we also include the results for the baseline model based on Support Vector Regression (SVR). We use SVR as the baseline method for PSPI estimation, as done in \cite{Kaltwang2012b}, and its personalized version, SVR-HCRF, using the proposed approach. Furthermore, to show the benefits of our hierarchical learning, we report the results obtained when the VAS estimation is attempted directly from input of the facial landmarks HCRF($X$)\, thus, skipping the PSPI estimation. We also show the results for estimation of VAS directly from the ground-truth PSPI. Note that we do not compare here with state-of-the-art models for PSPI estimation, since in this work we focus on the evaluation of VAS estimation, which has not been reported before in the literature. However, these approaches can be explored for replacing the first step of our method as future work.

{\bf Results.} From Fig.\ref{fig:confmatrixPSPI}, we also observe that PSPI estimated by RNN-LSTMs achieves accurate prediction of low intensity levels, while performing poorly on higher levels, likely due to the imbalance in the training data. However, Table \ref{fig:pspi_predictions} shows the improved PSPI estimation performance of the LSTM-RNNs over SVR~\cite{Kaltwang2012b} and non-recurrent NN in both measures. This is attributed to modeling of temporal information as well as different layers (more powerful feature extractors) by the former. These results are also similar to those obtained by \cite{unbc}, and can further be improved using some of the state-of-the-art methods for pain intensity estimation (e.g., \cite{Rodriguez2017,Egede2017}). Although the estimation of PSPI is not perfect when the base RNN/NN/SVR are used, they still can provide a good proxy of PSPI needed for estimation of VAS, as demonstrated below.

Table \ref{my-label} shows the VAS estimation results. The methods considered, from top down, are: (1) HCRF(PSPI) using manually coded PSPI values as input and HCRFs, (2) the proposed two-stage RNN-HCRF model (with automatically estimated PSPI), (3) the same as (2) but using SVR for the PSPI prediction (SVR+HCRF), and (4) HCRF($X$) using the facial landmarks (PCA pre-processed) as input. All four methods were evaluated with different personalization approaches, with $\alpha$ from 0 (no personalization), to 1 and 2 sequences of target persons (personalized), and with five random repetitions. Note that without the personalization, directly estimating VAS scores with HCRF($X$) outperforms the other models by a large margin. However, once the model personalization is performed, the proposed RNN-HCRF highly improves performance, outperforming both unpersonalized models ($\alpha=0$) as well as SVR+HCRF and HCRF($X$), while the latter two achieve similar performance. Finally, note that using only a single sequence of test persons to compute their I-FES, we attain a high boost in the models' performance.

To get more insights into the model's performance, in Fig.\ref{fig:subjects}, we show MAE\footnote{We do not report ICC here as it cannot be reliably computed per person because of small $\#$ of sequences.} for the VAS estimation per test person. Again, we vary the number of sequences to compute the I-FES for test subjects as: $\alpha=0,1,2$. Note that the proposed approach outperforms its unpersonalized counterpart ($\alpha=0$) on all test persons apart from 5\&21, and performs similarly on 13. A possible reason for this is that more sequences of these persons are needed to estimate their I-FES. On the other hand, note the high improvement on persons, e.g., 3,4,8 and 20. Finally, Fig.\ref{fig:confmatrixVAS} shows confusion matrices for the estimated VAS scores using the proposed pRNN-HCRF model. Observe the region in the upper left corner for the unpersonalized model ($\alpha=0$), where significant confusion occurs. These are corrected for by the proposed personalized approach.

\section{Conclusions}

Existing work on automatic pain estimation focuses on (arguably) objective pain measures such as PSPI, derived directly from facial action units. However, as we showed in this paper, for estimation of self-reported VAS, this traditional one-size-fits-all approach fails to account for individual differences. Yet, these proved to be critical for improving the estimation of VAS for each person. In this paper, we introduced and evaluated a new approach that can provide automatic personalized estimation of self-reported VAS pain levels. We showed, on the benchmark pain dataset of face images, that structuring the model to provide personalization results in improved estimation of VAS. The focus of our future research will be on investigating the robustness of the proposed expressiveness measure and on alternative measures that can be used to capture the individual biases inherently present in self-reports of pain. We also plan to extend our inference algorithm to be able to learn jointly the PSPI and VAS estimation. Finally, we plan to perform a more extensive evaluation of the proposed approach and investigate further the relationship between different pain scores (VAS, OPI, PSPI), and their automatic estimation from facial expressions.

\section*{Acknowledgements}
The work of O. Rudovic is funded by European Union H2020, Marie Curie Action - Individual Fellowship no. 701236 (EngageMe).


{\small
\bibliographystyle{ieee}
\bibliography{main}

\begin{thebibliography}{10}\itemsep=-1pt

\bibitem{Aasted2016}
C.~M. Aasted, M.~A. Yucel, S.~C. Steele, K.~Peng, D.~A. Boas, L.~Becerra, and
  D.~Borsook.
\newblock {Frontal lobe hemodynamic responses to painful stimulation: A
  potential brain marker of nociception}.
\newblock {\em PLoS ONE}, 11(11):1--12, 2016.

\bibitem{Aicher2012}
B.~Aicher, H.~Peil, B.~Peil, and H.-C. Diener.
\newblock {Pain measurement: Visual Analogue Scale (VAS) and Verbal Rating
  Scale (VRS) in clinical trials with OTC analgesics in headache.}
\newblock {\em Cephalalgia : an international journal of headache},
  32(3):185--97, 2012.

\bibitem{Ashraf2009a}
A.~B. Ashraf, S.~Lucey, J.~F. Cohn, T.~Chen, Z.~Ambadar, K.~M. Prkachin, and
  P.~E. Solomon.
\newblock {The painful face - Pain expression recognition using active
  appearance models}.
\newblock {\em Image and Vision Computing}, 27(12):1788--1796, 11 2009.

\bibitem{DeCWilliams2016}
A.~C.~de C~Williams and K.~D. Craig.
\newblock {Updating the definition of pain}.
\newblock {\em Pain}, Publish Ah(Box 1):1--14, 2016.

\bibitem{Calvo2010}
R.~A. Calvo and S.~D'Mello.
\newblock {Affect Detection: An Interdisciplinary Review of Models, Methods,
  and Their Applications}.
\newblock {\em IEEE Transactions on Affective Computing}, 1(1):18--37, 1 2010.

\bibitem{weixuan2017}
W.~Chen, A.~Sano, D.~Lopez~Martinez, S.~Taylor, A.~W. Mchill, A.~J.~K.
  Phillips, L.~Barger, E.~B. Klerman, and R.~W. Picard.
\newblock {Multimodal Ambulatory Sleep Detection}.
\newblock {\em IEEE International Conference on Biomedical and Health
  Informatics (BHI)}, pages 465--468, 2017.

\bibitem{Calvo2015}
J.~F. Cohn and F.~De~La~Torre.
\newblock {Automated Face Analysis for Affective Computing}.
\newblock In R.~Calvo, S.~D'Mello, J.~Gratch, and A.~Kappas, editors, {\em The
  Oxford Handbook of Affective Computing}, number June 2017, pages 1--37.
  Oxford University Press, 1 2015.

\bibitem{Egede2017}
J.~Egede, M.~Valstar, and B.~Martinez.
\newblock {Fusing Deep Learned and Hand-Crafted Features of Appearance, Shape,
  and Dynamics for Automatic Pain Estimation}.
\newblock 1 2017.

\bibitem{Ekman2002a}
P.~Ekman and W.~Friesen.
\newblock {Facial action coding system}.
\newblock 2002.

\bibitem{Farrar2000}
J.~T. Farrar, J.~T. Farrar, R.~K. Portenoy, R.~K. Portenoy, J.~a. Berlin, J.~a.
  Berlin, J.~L. Kinman, J.~L. Kinman, B.~L. Strom, and B.~L. Strom.
\newblock {Defining the clinically important difference in pain outcome
  measures}.
\newblock {\em Pain}, 88:287--294, 2000.

\bibitem{Ferreira-Valente2011}
M.~A. Ferreira-Valente, J.~L. Pais-Ribeiro, and M.~P. Jensen.
\newblock {Validity of four pain intensity rating scales}.
\newblock {\em Pain}, 152(10):2399--2404, 2011.

\bibitem{Graves2014}
A.~Graves and N.~Jaitly.
\newblock {Towards End-to-End Speech Recognition with Recurrent Neural
  Networks}.
\newblock In {\em Proc. 31st Int. Conf. Mach. Learn.}, volume~32, page
  1764–1772, 2014.

\bibitem{Gunes2009}
H.~Gunes and M.~Piccardi.
\newblock {Automatic Temporal Segment Detection and Affect Recognition From
  Face and Body Display}.
\newblock {\em IEEE Transactions on Systems, Man, and Cybernetics, Part B
  (Cybernetics)}, 39(1):64--84, 2 2009.

\bibitem{Haefeli2006}
M.~Haefeli and A.~Elfering.
\newblock {Pain assessment}.
\newblock {\em European Spine Journal}, 15(S1):S17--S24, 1 2006.

\bibitem{Hammal2012a}
Z.~Hammal and J.~F. Cohn.
\newblock {Automatic detection of pain intensity}.
\newblock In {\em Proceedings of the 14th ACM international conference on
  Multimodal interaction - ICMI '12}, page~47, New York, New York, USA, 2012.
  ACM Press.

\bibitem{Hawker2011}
G.~A. Hawker, S.~Mian, T.~Kendzerska, and M.~French.
\newblock {Measures of adult pain: Visual Analog Scale for Pain (VAS Pain),
  Numeric Rating Scale for Pain (NRS Pain), McGill Pain Questionnaire (MPQ),
  Short-Form McGill Pain Questionnaire (SF-MPQ), Chronic Pain Grade Scale
  (CPGS), Short Form-36 Bodily Pain Scale (SF}.
\newblock {\em Arthritis Care and Research}, 63(SUPPL. 11):240--252, 2011.

\bibitem{Hochreiter2001}
S.~Hochreiter, Y.~Bengio, P.~Frasconi, and J.~Schmidhuber.
\newblock {Gradient Flow in Recurrent Nets: the Difficulty of Learning
  Long-Term Dependencies}.
\newblock 2001.

\bibitem{Hochreiter1997}
S.~Hochreiter and J.~Schmidhuber.
\newblock {Long Short-Term Memory}.
\newblock {\em Neural computation}, 9(8):1735--80, 1997.

\bibitem{Jensen2003}
M.~P. Jensen, C.~Chen, and A.~M. Brugger.
\newblock {Interpretation of visual analog scale ratings and change scores: A
  reanalysis of two clinical trials of postoperative pain}.
\newblock {\em Journal of Pain}, 4(7):407--414, 2003.

\bibitem{Jensen2005}
M.~P. Jensen, S.~A. Martin, and R.~Cheung.
\newblock {The meaning of pain relief in a clinical trial}.
\newblock {\em Journal of Pain}, 6(6):400--406, 2005.

\bibitem{Joshi2013}
J.~Joshi, R.~Goecke, G.~Parker, and M.~Breakspear.
\newblock {Can body expressions contribute to automatic depression analysis?}
\newblock In {\em 2013 10th IEEE International Conference and Workshops on
  Automatic Face and Gesture Recognition (FG)}, pages 1--7. IEEE, 4 2013.

\bibitem{Kaltwang2012b}
S.~Kaltwang, O.~Rudovic, and M.~Pantic.
\newblock {Continuous Pain Intensity Estimation from Facial Expressions}.
\newblock In {\em Advances in Visual Computing}, pages 368--377. 2012.

\bibitem{Kim2010}
M.~Kim and V.~Pavlovic.
\newblock {Structured Output Ordinal Regression for Dynamic Facial Emotion
  Intensity Prediction}.
\newblock In {\em Lecture Notes in Computer Science (including subseries
  Lecture Notes in Artificial Intelligence and Lecture Notes in
  Bioinformatics)}, volume 6313 LNCS, pages 649--662. 2010.

\bibitem{Lafferty2001b}
J.~Lafferty, A.~McCallum, and F.~C.~N. Pereira.
\newblock {Conditional random fields: Probabilistic models for segmenting and
  labeling sequence data}.
\newblock {\em ICML '01 Proceedings of the Eighteenth International Conference
  on Machine Learning}, 8(June):282--289, 2001.

\bibitem{Lucey2011}
P.~Lucey, J.~F. Cohn, I.~Matthews, S.~Lucey, S.~Sridharan, J.~Howlett, and
  K.~M. Prkachin.
\newblock {Automatically detecting pain in video through facial action units}.
\newblock {\em IEEE Transactions on Systems, Man, and Cybernetics, Part B:
  Cybernetics}, 41(3):664--674, 2011.

\bibitem{unbc}
P.~Lucey, J.~F. Cohn, K.~M. Prkachin, P.~E. Solomon, and I.~Matthews.
\newblock {Painful data: The UNBC-McMaster shoulder pain expression archive
  database}.
\newblock {\em 2011 IEEE International Conference on Automatic Face and Gesture
  Recognition and Workshops, FG 2011}, pages 57--64, 2011.

\bibitem{Mackey2013}
S.~C. Mackey.
\newblock {Central neuroimaging of pain}.
\newblock {\em Journal of Pain}, 14(4):328--331, 2013.

\bibitem{textbookPain21}
J.~Melzack, Ronald;~Katz.
\newblock {Pain Measurement in Adult Patients}.
\newblock In S.~McMahon, M.~Koltzenburg, I.~Tracey, and D.~C. Turk, editors,
  {\em Wall {\&} Melzack's Textbook of Pain}, chapter~21. Elsevier, 6th
  edition, 2013.

\bibitem{Mohan2010}
H.~Mohan, J.~Ryan, B.~Whelan, and A.~Wakai.
\newblock {The end of the line? The Visual Analogue Scale and Verbal Numerical
  Rating Scale as pain assessment tools in the emergency department.}
\newblock {\em Emergency medicine journal : EMJ}, 27(5):372--375, 2010.

\bibitem{Monwar2006}
M.~Monwar and S.~Rezaei.
\newblock {Pain Recognition Using Artificial Neural Network}.
\newblock In {\em 2006 IEEE International Symposium on Signal Processing and
  Information Technology}, pages 28--33. IEEE, 8 2006.

\bibitem{Moore1997}
A.~Moore, O.~Moore, H.~McQuay, and D.~Gavaghan.
\newblock {Deriving dichotomous outcome measures from continuous data in
  randomised controlled trials of analgesics: Use of pain intensity and visual
  analogue scales}.
\newblock {\em Pain}, 69(3):311--315, 1997.

\bibitem{Price1983}
D.~D. Price, P.~A. McGrath, A.~Rafii, and B.~Buckingham.
\newblock {The validation of visual analogue scales as ratio scale measures for
  chronic and experimental pain}.
\newblock {\em Pain}, 17(1):45--56, 1983.

\bibitem{Prkachin2008}
K.~M. Prkachin and P.~E. Solomon.
\newblock {The structure, reliability and validity of pain expression: Evidence
  from patients with shoulder pain}.
\newblock {\em Pain}, 139(2):267--274, 2008.

\bibitem{Quattoni2004}
A.~Quattoni, M.~Collins, and T.~Darrell.
\newblock {Conditional Random Fields for Object Recognition}.
\newblock {\em Advances in Neural Information Processing Systems}, pages
  1097--1104, 2004.

\bibitem{Rodriguez2017}
P.~Rodriguez, G.~Cucurull, J.~Gonalez, J.~M. Gonfaus, K.~Nasrollahi, T.~B.
  Moeslund, and F.~X. Roca.
\newblock {Deep Pain: Exploiting Long Short-Term Memory Networks for Facial
  Expression Classification}.
\newblock {\em IEEE Transactions on Cybernetics}, pages 1--11, 2017.

\bibitem{Rudovic2013}
O.~Rudovic, V.~Pavlovic, and M.~Pantic.
\newblock {Automatic Pain Intensity Estimation with Heteroscedastic Conditional
  Ordinal Random Fields}.
\newblock In {\em Lecture Notes in Computer Science (including subseries
  Lecture Notes in Artificial Intelligence and Lecture Notes in
  Bioinformatics)}, volume 8034 LNCS, pages 234--243. 2013.

\bibitem{RudovicEtAlPAMI14}
O.~Rudovic, V.~Pavlovic, and M.~Pantic.
\newblock {Context-Sensitive Dynamic Ordinal Regression for Intensity
  Estimation of Facial Action Units}.
\newblock {\em IEEE Transactions on Pattern Analysis and Machine Intelligence},
  37(5):944--958, 5 2015.

\bibitem{Rusk2016}
A.~Ruiz, O.~Rudovic, X.~Binefa, and M.~Pantic.
\newblock {Multi-Instance Dynamic Ordinal Random Fields for Weakly-Supervised
  Pain Intensity Estimation}.
\newblock {\em Asian Conference on Computer Vision}, 9905(1):35--35, 2016.

\bibitem{Shrout1979}
P.~E. Shrout and J.~L. Fleiss.
\newblock {Intraclass correlations: Uses in assessing rater reliability.}
\newblock {\em Psychological Bulletin}, 86(2):420--428, 1979.

\bibitem{Sikka2013}
K.~Sikka, A.~Dhall, and M.~Bartlett.
\newblock {Weakly supervised pain localization using multiple instance
  learning}.
\newblock In {\em 2013 10th IEEE International Conference and Workshops on
  Automatic Face and Gesture Recognition (FG)}. IEEE, 4 2013.

\bibitem{Sullivan2000}
M.~J.~L. Sullivan, D.~A. Tripp, and D.~Santor.
\newblock {Gender differences in pain and pain behavior: The role of
  catastrophizing}.
\newblock {\em Cognitive Therapy and Research}, 24(1):121--134, 2000.

\bibitem{Wang2006a}
{Sy Bor Wang}, A.~Quattoni, L.-P. Morency, D.~Demirdjian, and T.~Darrell.
\newblock {Hidden Conditional Random Fields for Gesture Recognition}.
\newblock In {\em 2006 IEEE Computer Society Conference on Computer Vision and
  Pattern Recognition - Volume 2 (CVPR'06)}, volume~2, pages 1521--1527. IEEE,
  2006.

\bibitem{Todd1996}
K.~H. Todd.
\newblock {Clinical versus statistical significance in the assessment of pain
  relief}.
\newblock {\em Annals of Emergency Medicine}, 27(4):439--441, 1996.

\bibitem{Treister2012}
R.~Treister, M.~Kliger, G.~Zuckerman, I.~G. Aryeh, and E.~Eisenberg.
\newblock {Differentiating between heat pain intensities: The combined effect
  of multiple autonomic parameters}.
\newblock {\em Pain}, 153(9):1807--1814, 2012.

\bibitem{Wager2013}
T.~D. Wager, L.~Y. Atlas, M.~a. Lindquist, M.~Roy, C.-W. Woo, and E.~Kross.
\newblock {An fMRI-based neurologic signature of physical pain.}
\newblock {\em The New England journal of medicine}, 368(15):1388--97, 2013.

\bibitem{werner2012cl}
P.~Werner, A.~Al-Hamadi, and R.~Niese.
\newblock {Pain recognition and intensity rating based on Comparative
  Learning}.
\newblock In {\em 2012 19th IEEE International Conference on Image Processing},
  pages 2313--2316. IEEE, 9 2012.

\bibitem{Wilkie95}
D.~J. Wilkie.
\newblock {Facial Expressions of Pain in Lung Cancer}.
\newblock {\em Analgesia}, 1(2):91--99, 1995.

\bibitem{Williams2002a}
A.~C. D.~C. Williams.
\newblock {Facial expression of pain: an evolutionary account.}
\newblock {\em The Behavioral and brain sciences}, 25(4):439--455, 2002.

\bibitem{Williamson2005}
A.~Williamson and B.~Hoggart.
\newblock {Pain: A review of three commonly used pain rating scales}.
\newblock {\em Journal of Clinical Nursing}, 14(7):798--804, 2005.

\bibitem{Younger2009}
J.~Younger, R.~McCue, and S.~Mackey.
\newblock {Pain outcomes: A brief review of instruments and techniques}.
\newblock {\em Current Pain and Headache Reports}, 13(1):39--43, 2009.

\bibitem{Yucel2015}
M.~A. Y{\"{u}}cel, C.~M. Aasted, M.~P. Petkov, D.~Borsook, D.~A. Boas, and
  L.~Becerra.
\newblock {Specificity of hemodynamic brain responses to painful stimuli: a
  functional near-infrared spectroscopy study.}
\newblock {\em Scientific reports}, 5:9469, 2015.

\bibitem{zeng2009survey}
{Zhihong Zeng}, M.~Pantic, G.~Roisman, and T.~Huang.
\newblock {A Survey of Affect Recognition Methods: Audio, Visual, and
  Spontaneous Expressions}.
\newblock {\em IEEE Transactions on Pattern Analysis and Machine Intelligence},
  31(1):39--58, 1 2009.

\end{thebibliography}
}

\end{document}